\documentclass[sigconf]{acmart}
\usepackage{xcolor}         
\usepackage{xspace}
\usepackage{graphicx}
\usepackage{float}

\usepackage{booktabs}

\usepackage{paralist}
\usepackage{wrapfig}
\usepackage{amsmath} 
\usepackage{threeparttable}
\usepackage{multirow}
\usepackage{algorithm}
\usepackage{algpseudocode}
\usepackage{tabularx}

\usepackage{diagbox}

\newcommand{\model}{\textbf{\textsc{HyGRAG}}\xspace} 
 
\newcommand{\vpara}[1]{\vspace{0.05in}\noindent\textbf{#1 }}
\AtBeginDocument{%
  }

\copyrightyear{2026}
\acmYear{2026}
\setcopyright{cc}
\setcctype{by}
\acmConference[WWW '26]{Proceedings of the ACM Web Conference 2026}{April 13--17, 2026}{Dubai, United Arab Emirates}
\acmBooktitle{Proceedings of the ACM Web Conference 2026 (WWW '26), April 13--17, 2026, Dubai, United Arab Emirates}
\acmPrice{}
\acmDOI{10.1145/3774904.3792720}
\acmISBN{979-8-4007-2307-0/2026/04}

\begin{document}

\title{A Unified Framework for Context-Aware and Relation-Aware Graph Retrieval-Augmented Generation}

\author{Haoyang Zhong}\authornote{Work done while visiting Zhejiang University.}
\affiliation{%
  \institution{Zhejiang University}
  \city{Hangzhou}
  \country{China}
}

\affiliation{%
  \institution{Nanyang Technological University}
  \city{Singapore}
  \country{Singapore}
}
\email{haoyang011@e.ntu.edu.sg}

\author{Yifei Sun}
\authornote{Corresponding author.}
\affiliation{%
  \institution{Zhejiang University}
  \city{Hangzhou}
  \country{China}}
\email{yifeisun@zju.edu.cn}

\author{Antong Zhang}
\affiliation{%
  \institution{Zhejiang University}
  \city{Hangzhou}
  \country{China}}
\email{antongzhang@zju.edu.cn}

\author{Chunping Wang}
\affiliation{%
 \institution{Finvolution Group}
 \city{Shang Hai}
 \country{China}}
\email{wangchunping02@xinye.com}

\author{Lei Chen}
\affiliation{%
 \institution{Finvolution Group}
 \city{Shang Hai}
 \country{China}}
\email{chenlei04@xinye.com}

\author{Yang Yang}
\affiliation{%
  \institution{Zhejiang University}
  \city{Hangzhou}
  \country{China}}
\email{yangya@zju.edu.cn}

\renewcommand{\shortauthors}{Haoyang Zhong et al.}

\begin{abstract}
Retrieval-Augmented Generation (RAG) has emerged as a paradigm for enhancing large language models (LLMs) with external knowledge, yet existing graph-based methods face a fundamental limitation: entity-centric and chunk-centric approaches operate on representations anchored to original text without true knowledge fusion. While entity-centric methods connect logically related content and chunk-centric methods preserve context, both retrieve information separately through similarity search, missing emergent understanding from their synthesis. In this paper, we propose \model, a hierarchical graph RAG framework that transcends source documents by addressing three core challenges: constructing summaries that genuinely integrate contextual and relational information, leveraging these synthesized representations to access emergent knowledge during retrieval, and efficiently updating hierarchical structures for dynamic corpora. Specifically, we design hierarchical index structures over hybrid graphs with both chunk and entity nodes, then iteratively cluster them and generate LLM-based summaries. Then, we design context and relation-aware retrieval that searches across all abstraction levels while expanding through community membership. Moreover, we enable dynamic knowledge update through attachment-based algorithms with only local re-summarization. Experimental results show that \model improves the average accuracy of multi-hop reasoning tasks by 9.7\%, while maintaining reasonable efficiency. 
\footnote[1]{Our codes are available at \url{https://github.com/zjunet/HyGRAG}.}
\end{abstract}

\ccsdesc[500]{Networks~Online social networks}
\ccsdesc[300]{Computing methodologies~Knowledge representation and reasoning}

\keywords{Graph RAG; context-aware; relation-aware.}


\maketitle

\section{Introduction}

\begin{figure}[t!]
    \centering
    \includegraphics[width=0.48\textwidth]{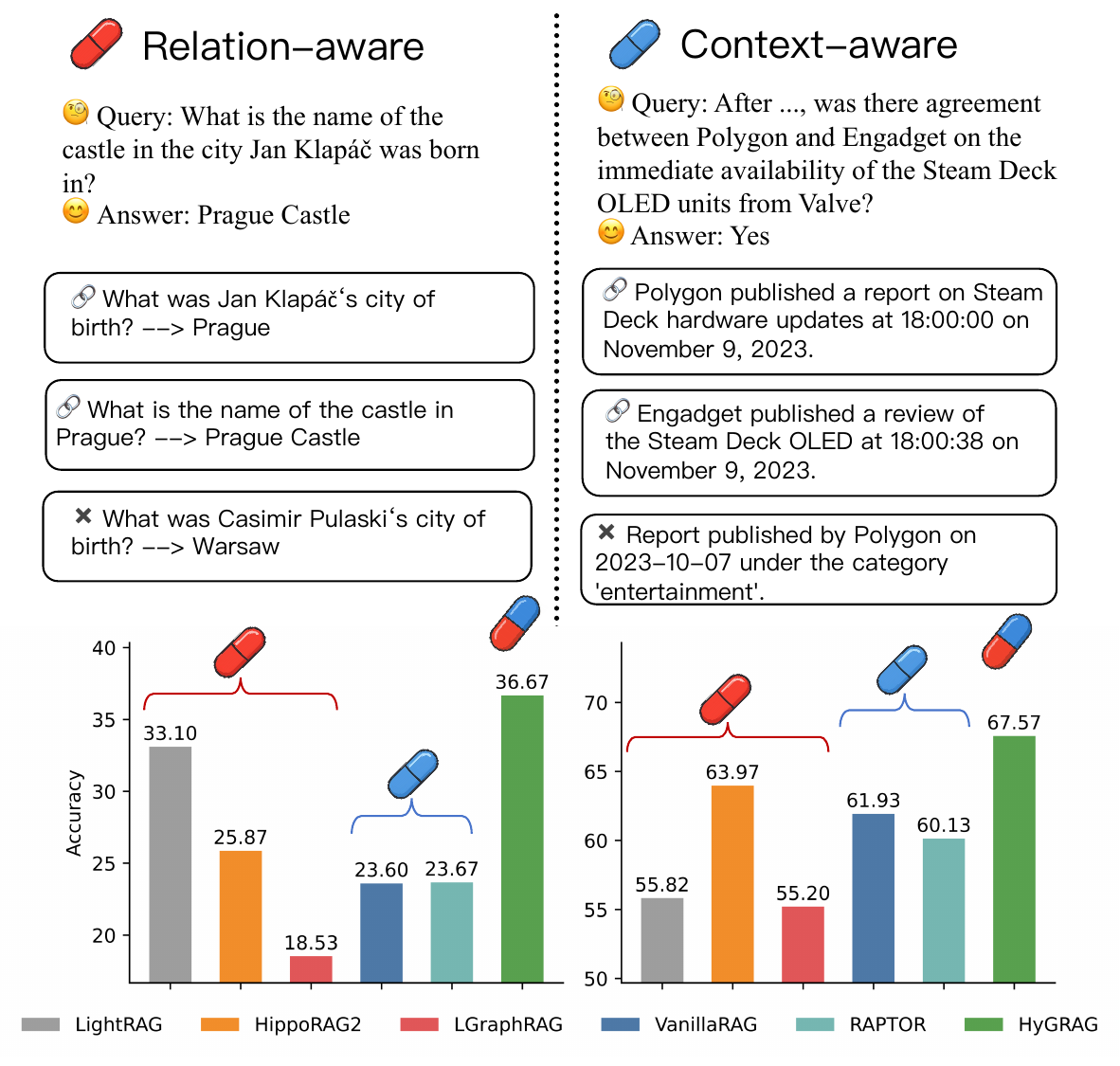}
    \setlength{\abovecaptionskip}{-0.3cm}   
    \setlength{\belowcaptionskip}{-0.5cm} 
    \caption{Illustration of method performance using Qwen3-8B across relation-aware (MuSiQue) and context-aware (MultiHop-RAG) datasets, including representative cases.}
    \label{fig:obs}
\end{figure}

RAG systems have been developed to enhance LLMs by integrating external knowledge sources \cite{gao2024retrievalaugmentedgenerationlargelanguage,lewis2021retrievalaugmentedgenerationknowledgeintensivenlp}. This integration allows LLMs to access up-to-date information and domain-specific knowledge, addressing the inherent limitations of parametric memory \cite{fan2024surveyragmeetingllms}. Recent advances in RAG have moved beyond simple document retrieval to incorporate graph structures that explicitly model relationships between concepts, entities, and documents \cite{hu2025graggraphretrievalaugmentedgeneration,xiang2025usegraphsragcomprehensive}. These graph RAG systems leverage the rich structural information in knowledge graphs to improve both retrieval accuracy and reasoning capabilities, making them particularly effective for complex queries that require understanding interconnected information \cite{ma2025thinkongraph20deepfaithful,10.5555/3737916.3739818}.

Current Graph RAG methods have explored two distinct directions to harness graph structures for retrieval augmentation. Entity-centric approaches such as GraphRAG \cite{edge2025localglobalGraphRAG}, HippoRAG \cite{10.5555/3737916.3739818}, and HiRAG \cite{huang2025retrievalaugmentedgenerationhierarchicalknowledge} build knowledge graphs where nodes represent entities extracted from text and edges encode their semantic relationships. These methods excel at multi-hop reasoning by enabling traversal along relational paths—for instance, answering "Which company acquired the developer of GPT-3?" by following edges from GPT-3 to OpenAI to Microsoft. In contrast, chunk-centric methods like RAPTOR \cite{sarthi2024raptor} and EraRAG \cite{zhang2025eraragefficientincrementalretrieval} organize text chunks into hierarchical structures based on semantic similarity. They create tree-like indexes where leaf nodes contain original text chunks and parent nodes store increasingly abstract summaries. This design preserves full contextual information while enabling retrieval at different levels of granularity.

However, both approaches exhibit fundamental limitations that restrict their effectiveness. Entity-centric methods suffer from information loss during the entity extraction process. Named entity recognition and relation extraction models introduce errors that compound throughout the system, and the abstraction from text to entities discards valuable contextual information needed for factual question answering. Our analysis shows that these methods often underperform simple dense retrieval on factual QA benchmarks. Chunk-centric methods face the opposite problem: while they maintain high accuracy on factual queries, they cannot capture explicit relationships between entities scattered across different chunks. This makes them ineffective for queries requiring logical reasoning, such as finding all products affected by a specific supply chain disruption. The fundamental issue is that existing methods optimize for either contextual completeness or relational structure, but real-world queries often require both.


Our key insight is that simply combining chunk and entity representations in a hybrid graph does not solve the fundamental problem: both representations remain anchored to their original textual sources without true knowledge fusion. While knowledge graphs can connect logically related content that is textually distant, they cannot integrate the logical information from relations with the background context from chunks. During retrieval, these two aspects are still accessed separately through similarity search, missing the opportunity for synergistic understanding. Consider a query about "the impact of renewable energy adoption on manufacturing industries"—entity-based retrieval might find relationships between solar panels and factories, while chunk-based retrieval might return documents about energy costs, but neither captures the emergent understanding that comes from synthesizing these perspectives. This motivates our approach: we cluster entities and chunks together and generate summaries that simultaneously encode both relational logic and contextual background, creating new knowledge representations that transcend the original text.

Implementing this knowledge fusion approach presents three core technical challenges. First, how to construct summaries that genuinely integrate context and relations rather than simply concatenating them. Second, how to leverage these hybrid summaries during retrieval to access knowledge beyond the original corpus. The summaries represent emergent understanding that doesn't exist in any single source document—the retrieval mechanism must be able to match queries to these synthesized insights while still providing access to supporting details. Third, how to efficiently update such complex hierarchical structures when the corpus changes. Real-world knowledge bases receive continuous updates, but rebuilding the entire hierarchy for each new document would be computationally prohibitive.

To address these challenges, we propose \model, a unified framework with three key innovations. For summary construction, we design a two-stage approach: first clustering nodes using graph structure-aware embeddings that capture both textual similarity and relational connectivity, then prompting LLMs with structured templates that explicitly request synthesis of contextual and relational information within each community. This produces summaries that represent genuine knowledge fusion rather than simple aggregation. For retrieval, we implement a bi-level mechanism that operates across both context and relation dimensions. Context-aware retrieval searches across all hierarchy levels—from specific chunks to abstract community summaries—enabling access to emergent knowledge. Relation-aware retrieval then expands results by collecting entities from retrieved communities and filtering their associated triplets, ensuring logical completeness. For dynamic updates, we develop an attachment-based algorithm where new content is matched to the most similar community at the appropriate abstraction level, with local re-summarization propagating only along affected paths. This preserves the stability of unrelated portions while incorporating new knowledge. Experimental results show that \model improves the average accuracy of multi-hop reasoning tasks by 9.7\%, 
while maintaining reasonable efficiency.

In summary, we make three contributions:
\begin{itemize}
\item We propose hierarchical indexing for Graph RAG that unifies context and relational information through multi-level abstraction, enabling retrieval across different granularities.
\item We develop a retrieval mechanism that combines context-aware and relation-aware strategies, leveraging community structures to find information that flat approaches miss.
\item We achieve significant performance gains across a range of tasks, including a 6.2\% improvement in Factual Accuracy and a 9.7\% increase in Multi-Hop Reasoning, with improvements reaching up to 12.2\% on the HotpotQA dataset.
\end{itemize}
\section{Related Work}

\subsection{Graph RAG}

RAG~\cite{gao2024retrievalaugmentedgenerationlargelanguage} mitigates hallucinations~\cite{Huang_2025} in LLMs by integrating external knowledge during generation. RAG involves two coupled steps: retrieval and generation. Given a query, the system retrieves the top-k relevant text chunks from a preprocessed external corpus. The LLM then generates an answer conditioned on both the retrieved chunks and the query, producing more factual and contextually appropriate responses.

Vanilla RAG~\cite{10.5555/3495724.3496517} improves factual grounding but treats the corpus as unstructured text, limiting inter-document reasoning and multi-hop inference. To overcome this, recent work incorporates graph structures into retrieval~\cite{peng2024graphretrievalaugmentedgenerationsurvey}, giving rise to graph RAG methods. These can be categorized as context-aware, focusing on text chunks, and relation-aware, explicitly leveraging graph structures.

\textbf{Context-aware} methods enhance generation by retrieving segmented text chunks as context. Vanilla RAG retrieves top-k chunks based on embedding similarity. Advanced approaches cluster and summarize chunks to capture high-order information. RAPTOR~\cite{sarthi2024raptor} recursively clusters chunks into hierarchical summaries, while EraRAG~\cite{zhang2025eraragefficientincrementalretrieval} organizes the corpus into multi-layered summaries using LSH, handling high-order information in dynamic corpora.

\textbf{Relation-aware} methods construct textual knowledge graphs to encode structured relationships. LightRAG~\cite{guo2025lightragsimplefastretrievalaugmented} integrates graph structures into the text indexing ,and retrieve by high/low-level keywords. GraphRAG~\cite{edge2025localglobalGraphRAG} cluster communities and summarize them to provide holistic context. HiRAG~\cite{huang2025retrievalaugmentedgenerationhierarchicalknowledge} and ArchRAG~\cite{wang2025archragattributedcommunitybasedhierarchical} further leverage hierarchical graphs to bridge local entity details and global insights by applying attributed entities. HippoRAG~\cite{10.5555/3737916.3739818} and HippoRAG 2~\cite{gutiérrez2025ragmemorynonparametriccontinual} enhance multi-hop reasoning by navigating graph connectivity via Personalized PageRank, improving performance on complex QA tasks.

\subsection{Dynamic Retrieval}

Most graph RAG systems assume a static corpus, making updates costly. LightRAG~\cite{guo2025lightragsimplefastretrievalaugmented} uses a modular retriever to add new documents dynamically. HippoRAG~\cite{10.5555/3737916.3739818} also supports incremental updates, as it leverages the KG only for retrieval. EraRAG~\cite{zhang2025eraragefficientincrementalretrieval} introduces LSH-based clustering, enabling localized updates by re-segmenting and re-summarizing only affected parts. However, entity-based community clustering methods still require full reconstruction.

\begin{figure*}[t!]
    \centering
    \includegraphics[width=0.85\textwidth]{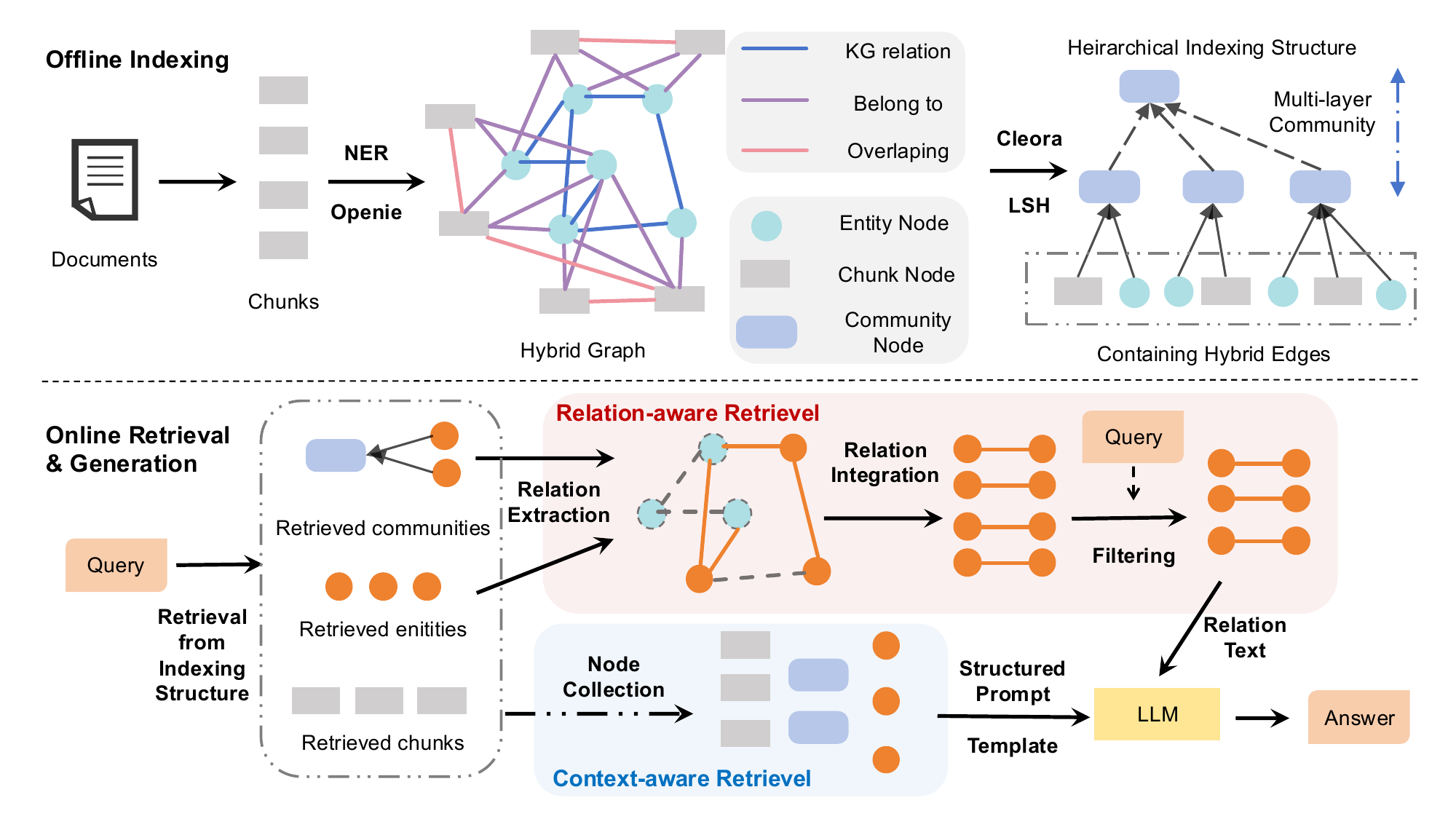}
    \setlength{\abovecaptionskip}{-0.05cm} 
    \setlength{\belowcaptionskip}{-0.23cm} 
    \caption{Overall architecture of \model. }
    \label{fig:model}
\end{figure*}

\section{Method}

\vpara{Overview.}
Our proposed framework, \model, addresses the fundamental trade-off between context-aware and relation-aware retrieval through a unified architecture. The system consists of four key components: (1) \textbf{Hierarchical Index Structure Construction} constructs a hierarchical indexing structure with hybrid graph that captures both chunk-based contextual information and entity-based relational information; (2) \textbf{Context and Relation-Aware Retrieval} performs retrieval across micro and macro granularities; (3) \textbf{Retrieval-Augmented Efficient Generation} ensemble heirarchical context; (4) \textbf{Dynamic Knowledge Update} supports incremental knowledge integration without full reconstruction. 

\subsection{Hierarchical Index Structure Construction}

This module operates entirely in the offline phase, ensuring no impact on online retrieval efficiency. We construct a hybrid knowledge graph from raw text that simultaneously captures chunk-based contextual information and entity-based relational structures.

\vpara{Text Chunking and Chunk-Level Graph Construction.}
Given a corpus $\mathcal{D}$, we first segment it into overlapping text chunks $\mathcal{C} = \{c_1, c_2, \ldots, c_n\}$. Each chunk $c_i$ preserves the contextual information of the original text, enabling inference LLMs to better understand its semantic content. For each chunk, we compute its embedding representation:
\begin{equation}
\mathbf{e}_i^c = \text{LM}_\text{Embedding}(c_i),
\end{equation}
where we employ BGE-M3 as the embedding model to encode text chunks into dense vector representations.
Based on embedding similarity, we construct a chunk-based graph $\mathcal{G}^c = (\mathcal{V}^c, \mathcal{E}^c)$, where $\mathcal{V}^c = \mathcal{C}$. 
We create edges between chunks based on shared entities rather than shallow embedding similarity. Specifically, if two chunks $c_i$ and $c_j$ share more than $l$ common entities, we create an edge $(c_i, c_j) \in \mathcal{E}^c$:
\begin{equation}
(c_i, c_j) \in \mathcal{E}^c \Leftrightarrow |\mathcal{E}_i \cap \mathcal{E}_j| > l,
\end{equation}
where $\mathcal{E}_i$ and $\mathcal{E}_j$ are the entity sets extracted from chunks $c_i$ and $c_j$ respectively, and $l$ is set to 3 in our implementation. This approach establishes meaningful connections between chunks through shared entities, providing stronger semantic relationships than surface-level embedding similarity.

\vpara{Entity-Level Graph Construction.}
Then we employ LLMs to extract knowledge graph triplets from the raw text. Specifically, for each text chunk $c_i$, the LLM identifies entities and their relationships, generating a set of triplets:
\begin{equation}
\mathcal{T}_i = \{(h, r, t) \mid h, t \in \mathcal{E}_i, r \in \mathcal{R}_i\},
\end{equation}
where $\mathcal{E}_i$ and $\mathcal{R}_i$ are the entity and relation sets extracted from $c_i$, respectively. By merging triplets from all chunks, we obtain the global entity set $\mathcal{V}^e = \bigcup_{i=1}^{n} \mathcal{E}_i$ and relation set $\mathcal{E}^e = \bigcup_{i=1}^{n} \mathcal{T}_i$, forming the entity-based knowledge graph $\mathcal{G}^e = (\mathcal{V}^e, \mathcal{E}^e)$.

\vpara{Hybrid Graph Construction.}
To bridge the gap between chunk-based context and entity-based relations, we connect the two graphs through membership relationships. For each entity $e \in \mathcal{V}^e$, we identify the set of text chunks containing that entity:
\begin{equation}
\mathcal{C}(e) = \{c_i \in \mathcal{C} \mid e \text{ is mentioned in } c_i\}.
\end{equation}
We create cross-layer edges $\mathcal{E}^{ce}$ connecting entities to their containing chunks:
\begin{equation}
\mathcal{E}^{ce} = \{(e, c) \mid e \in \mathcal{V}^e, c \in \mathcal{C}(e)\}.
\end{equation}
The final hybrid knowledge graph is defined as:
\begin{equation}
\mathcal{G}_{\text{hybrid}} = (\mathcal{V}^c \cup \mathcal{V}^e, \mathcal{E}^c \cup \mathcal{E}^e \cup \mathcal{E}^{ce}).
\end{equation}
This graph contains two types of nodes (chunk nodes and entity nodes) and three types of edges (chunk-to-chunk edges, entity-to-entity edges, and cross-layer edges).

\vpara{Hierarchical Indexing.}
While the hybrid graph enriches information representation, it also increases the complexity of online retrieval. To efficiently retrieve both relevant contextual and relational information simultaneously, we propose to construct a hierarchical tree-structured index for the hybrid graph.

First, we use Cleora to generate structure-aware embeddings for all nodes $v \in \mathcal{V}^c \cup \mathcal{V}^e$ in the hybrid graph:
\begin{equation}
\mathbf{z}_v = \text{Cleora}(\mathcal{G}_{\text{hybrid}}, v).
\end{equation}

Then, we employ hyperplane-based Locality-Sensitive Hashing (LSH) for efficient clustering. For each node embedding $\mathbf{z}_v \in \mathbb{R}^d$, we randomly sample $k$ hyperplanes $\{\mathbf{h}_1, \ldots, \mathbf{h}_k\}$ and compute its hash code:
\begin{equation}
\text{hash}(\mathbf{z}_v) = [\text{sign}(\mathbf{z}_v \cdot \mathbf{h}_1), \ldots, \text{sign}(\mathbf{z}_v \cdot \mathbf{h}_k)]
\end{equation}
Nodes with similar hash codes (small Hamming distance) are assigned to the same bucket. For each bucket $B$, if its size satisfies $S_{\min} \leq |B| \leq S_{\max}$, it is retained; otherwise, split or merge operations are performed:
\begin{equation}
B_{\text{adjusted}} = 
\begin{cases}
\text{Split}(B) & \text{if } |B| > S_{\max} \\
\text{Merge}(B, B_{\text{neighbor}}) & \text{if } |B| < S_{\min} \\
B & \text{otherwise}
\end{cases}
\end{equation}

For each adjusted bucket (forming a community) $C$, we generate summary representations through a two-step process:
\begin{equation}
t_C = \text{LLM}_{\text{summarize}}(\{v \mid v \in C\}),
\end{equation}

\begin{equation}
s_C = \text{LM}_{\text{Embedding}}(t_C),
\end{equation}

where we utilize Llama3.1-8B-Instruct to generate concise textual summaries $t_C$ that capture the semantic essence of nodes within each community, and then employ BGE-M3 to encode these summaries into dense vector representations $s_C$.

We treat summary nodes as nodes in the next layer and recursively apply the above hashing, bucketing, and summarization process up to a predefined number of layers. Let $\mathcal{L}_\ell$ denote the node set at layer $\ell$, where $\mathcal{L}0 = \mathcal{V}^c \cup \mathcal{V}^e$ represents the leaf nodes. The hierarchical construction continues until the layer index $\ell$ reaches the preset maximum depth $L$, resulting in an $L$-layer hierarchical index tree:
\begin{equation}
\mathcal{T}{\text{index}} = {\mathcal{L}_0, \mathcal{L}_1, \ldots, \mathcal{L}_L}.
\end{equation}
This hierarchical tree serves as an efficient retrieval database during the dual-stage retrieval phase, where leaf nodes correspond to original nodes from the hybrid graph and upper-layer nodes contain semantic summaries at coarser granularities, supporting multi-scale context and relation-aware retrieval.

\subsection{Context and Relation-Aware Retrieval}

During the online retrieval phase, we employ a bi-level approach that captures information at both context and relation levels to provide comprehensive and contextually rich results.

\vpara{Query Encoding.}
Given a user query $q$, we first encode it using the same embedding model employed during indexing:
\begin{equation}
\mathbf{q} = \text{LM}_{\text{Embedding}}(q),
\end{equation}
where we utilize BGE-M3 to generate the query embedding $\mathbf{q} \in \mathbb{R}^d$ that serves as the basis for similarity computation across all retrieval levels.

\vpara{Context-Aware Retrieval.}
From the hierarchical indexing structure $\mathcal{T}_{\text{index}}$ containing community, chunk, and entity nodes, we first perform context-aware retrieval by computing similarity scores across all node types. We retrieve the top-$k$ most similar nodes for each type:

\begin{equation}
\mathcal{P}_{\text{retrieved}} = \text{TopK}(\{C \in \bigcup_{\ell=1}^{L} \mathcal{L}_\ell \mid \text{sim}(\mathbf{q}, s_C)\}, k),
\end{equation}

\begin{equation}
\mathcal{C}_{\text{retrieved}} = \text{TopK}(\{c \in \mathcal{V}^c \mid \text{sim}(\mathbf{q}, \mathbf{e}_c^c)\}, k),
\end{equation}

\begin{equation}
\mathcal{E}_{\text{retrieved}}^{\text{context}} = \text{TopK}(\{e \in \mathcal{V}^e \mid \text{sim}(\mathbf{q}, \mathbf{z}_e)\}, k).
\end{equation}

These retrievals capture content most relevant to the query from a contextual perspective across different granularities.

\vpara{Relation-Aware Retrieval.}
To capture logical relationships, we construct a comprehensive entity set by combining entities from retrieved communities and context-based entity retrieval:
\begin{equation}
\mathcal{E}_{\text{all}} = \mathcal{E}_{\text{retrieved}}^{\text{context}} \cup \bigcup_{C \in \mathcal{P}_{\text{retrieved}}} \{e \mid e \in \mathcal{V}^e, (e, C) \in \mathcal{E}^{ce}\}.
\end{equation}

We then extract all relations connected to these entities from leaf nodes:
\begin{equation}
\mathcal{R}_{\text{all}} = \{(h, r, t) \in \mathcal{E}^e \mid h \in \mathcal{E}_{\text{all}} \vee t \in \mathcal{E}_{\text{all}}\}.
\end{equation}

Since $|\mathcal{R}_{\text{all}}|$ can be large, potentially introducing noise and LLM overhead, we filter by computing triplet embeddings and selecting the top-$k$ most relevant:
\begin{equation}
\mathcal{R}_{\text{retrieved}} = \text{TopK}(\{(h,r,t) \in \mathcal{R}_{\text{all}} \mid \mathbf{z}_{(h,r,t)} = \text{LM}_{\text{Embedding}}(h \oplus r \oplus t)\}, k),
\end{equation}
where $\oplus$ denotes concatenation.

\vpara{Integrated Retrieval Strategy.}
The final retrieval combines context-aware and relation-aware perspectives, yielding four sets of $k$ elements each:
\begin{equation}
\mathcal{R}_{\text{final}} = \{\mathcal{P}_{\text{retrieved}}, \mathcal{C}_{\text{retrieved}}, \mathcal{E}_{\text{retrieved}}^{\text{context}}, \mathcal{R}_{\text{retrieved}}\},
\end{equation}
providing community summaries for high-level understanding, chunks for detailed context, entities for key concepts, and relations for logical connections.

\vpara{Computational Complexity Analysis.}
Similar to other retrieval-augmented approaches, \model employs a FAISS vector store with HNSW indexing to efficiently retrieve the top-$k$ most similar items. Let $N$ denote the total number of indexed nodes and $d$ the embedding dimension. The HNSW-based search achieves logarithmic complexity for each index:$\mathcal{O}(\log N_i \cdot d),$
and retrieving from entity-, chunk-, and community-level indexes results in:
\begin{equation}
\mathcal{O}((\log N_e + \log N_c + \log N_p) \cdot d) = \mathcal{O}(\log N \cdot d).
\end{equation}
For each retrieved entity, relation extraction and relevance scoring introduce an additional cost of $\mathcal{O}(k_e \cdot \bar{d} \cdot d)$, where $\bar{d}$ is the average node degree. The overall retrieval complexity is thus:
\begin{equation}
\mathcal{O}(\log N \cdot d + k_e \cdot \bar{d} \cdot d),
\end{equation}
which maintains sub-linear scaling with respect to graph size while enabling multi-granularity and relation-aware retrieval.

Compared with traditional chunk-aware RAG systems that only perform a single chunk-level retrieval $\mathcal{O}(\log N_c \cdot d)$, our method introduces additional but lightweight costs from entity and community retrievals $\mathcal{O}((\log N_e + \log N_p) \cdot d + k_e \cdot \bar{d} \cdot d)$.

\subsection{Retrieval-Augmented Efficient Generation}

Given the retrieved nodes from bi-level retrieval, we extract and organize information for LLM generation. For community nodes $C \in \mathcal{P}_{\text{retrieved}}$, we obtain their summaries $t_C$. For entity nodes $e \in \mathcal{E}_{\text{retrieved}}^{\text{context}}$, we extract their textual representations. For relation triplets $(h,r,t) \in \mathcal{R}_{\text{retrieved}}$, we preserve their structured format. For chunk nodes $c \in \mathcal{C}_{\text{retrieved}}$, we directly use their textual content.

We combine these four information types using a structured prompt template:
\begin{equation}
\text{Context} = \mathcal{P}(\{t_C\}_{C \in \mathcal{P}_{\text{retrieved}}}, \mathcal{E}_{\text{retrieved}}^{\text{context}}, \mathcal{R}_{\text{retrieved}}, \mathcal{C}_{\text{retrieved}}),
\end{equation}
where $\mathcal{P}$ organizes community summaries, entities, relation triplets, and chunk contexts hierarchically.

The final response is generated by:
\begin{equation}
y = \text{LLM}_{\text{generate}}(q, \text{Context}),
\end{equation}
leveraging community summaries for high-level understanding, entities for key concepts, relation triplets for logical reasoning, and chunk contexts for detailed background information.

\subsection{Dynamic Knowledge Update}

In real-world scenarios, knowledge corpora evolve continuously, requiring efficient update mechanisms. Our clustering-based hierarchical structure enables fast integration of new content without full graph reconstruction.

\vpara{Update Processing.} Given new text content $d_{\text{new}}$, we first segment it into chunks $\mathcal{C}_{\text{new}} = \{c_{\text{new}}^1, \ldots, c_{\text{new}}^m\}$ and extract knowledge graph triplets:
\begin{equation}
\mathcal{T}_{\text{new}} = \bigcup_{i=1}^{m} \{(h, r, t) \mid h, t \in \mathcal{E}_i^{\text{new}}, r \in \mathcal{R}_i^{\text{new}}\}.
\end{equation}

We generate a summary representation for the new content:
\begin{equation}
t_{\text{new}} = \text{LLM}_{\text{summarize}}(\mathcal{C}_{\text{new}} \cup \mathcal{T}_{\text{new}}),
\end{equation}
\begin{equation}
\mathbf{s}_{\text{new}} = \text{LM}_{\text{Embedding}}(t_{\text{new}}).
\end{equation}

\vpara{Hierarchical Attachment.} We traverse the hierarchical index from bottom to top. Starting at layer $\mathcal{L}_1$, we find the most similar community:
\begin{equation}
C^* = \arg\max_{C \in \mathcal{L}_1} \text{sim}(\mathbf{s}_{\text{new}}, s_C).
\end{equation}

If $\text{sim}(\mathbf{s}_{\text{new}}, s_{C^*}) > \tau_{\text{attach}}$, we attach the new content to $C^*$. Otherwise, we proceed to the next layer:
\begin{equation}
\ell^* = \min\{\ell \mid \exists C \in \mathcal{L}_\ell : \text{sim}(\mathbf{s}_{\text{new}}, s_C) > \tau_{\text{attach}}\}.
\end{equation}

Upon attachment at layer $\ell^*$, we update all ancestor community summaries along the path to the root:
\begin{equation}
\forall j \in \{\ell^*, \ldots, L\}, \quad t_{C_j} \leftarrow \text{LLM}_{\text{summarize}}(\text{Children}(C_j)),
\end{equation}
\begin{equation}
s_{C_j} \leftarrow \text{LM}_{\text{Embedding}}(t_{C_j}).
\end{equation}

Additionally, we establish connections between new nodes and existing leaf nodes following the original hybrid graph construction rules, creating edges when entities are shared or semantic similarity exceeds thresholds. Since all updates occur offline, online retrieval efficiency remains unaffected.

\section{Experiments}

In this section, we answer the following questions to validate the effectiveness of our method:
\begin{itemize}
    \item \textbf{RQ1.} How does \model perform compared with existing baselines on different type of static QA tasks?
    \item \textbf{RQ2.} How efficient is the \model approach?
    \item \textbf{RQ3.} How robust is \model in corpus expansion?
    \item \textbf{RQ4.} What is the quality of our communities and relations, and how does it impact the performance of RAG?
\end{itemize}

\subsection{Experimental Setup}
\subsubsection{Datasets and Metrics}

For static QA, we adopt five datasets: \textbf{PopQA} \cite{mallen2023trustlanguagemodelsinvestigating} (factual accuracy), \textbf{MuSiQue} \cite{trivedi2022musiquemultihopquestionssinglehop}, and \textbf{HotpotQA} \cite{yang2018hotpotqadatasetdiverseexplainable} (multi-hop reasoning), and \textbf{MultiHop-RAG} \cite{tang2024multihopragbenchmarkingretrievalaugmentedgeneration}, \textbf{QuALITY} \cite{pang2022qualityquestionansweringlong} (reading comprehension).
For evaluation, static datasets report \textbf{Accuracy} and \textbf{Recall} (QuALITY uses Accuracy only).

\begin{table*}[ht]
\centering
    \setlength{\belowcaptionskip}{-0.05cm} 
\caption{Overall QA results(Accuracy and Recall) on static RAG query datasets using Llama-3.1-8B-Instruct. 
}
\label{tab:static_results}
\resizebox{0.8\textwidth}{!}{
\begin{tabular}{lccccccccc} 
\toprule
\textbf{Method} & \multicolumn{2}{c}{\textbf{Factual Accuracy}} & \multicolumn{3}{c}{\textbf{Reading Comprehension}} & \multicolumn{4}{c}{\textbf{Multi-Hop Reasoning}} \\ 
\cmidrule(lr){2-3} \cmidrule(lr){4-6} \cmidrule(lr){7-10} 
& \multicolumn{2}{c}{PopQA} & \multicolumn{1}{c}{QuALITY} & \multicolumn{2}{c}{MultiHop-RAG} & \multicolumn{2}{c}{MuSiQue} & \multicolumn{2}{c}{HotpotQA} \\ 
\cmidrule(lr){2-3} \cmidrule(lr){4-4} \cmidrule(lr){5-6} \cmidrule(lr){7-8} \cmidrule(lr){9-10} 
& \multicolumn{1}{c}{Accuracy} & \multicolumn{1}{c}{Recall} & \multicolumn{1}{c}{Accuracy} & \multicolumn{1}{c}{Accuracy} & \multicolumn{1}{c}{Recall} & \multicolumn{1}{c}{Accuracy} & \multicolumn{1}{c}{Recall} & \multicolumn{1}{c}{Accuracy} & \multicolumn{1}{c}{Recall} \\
\midrule
Zero-shot & 15.37 & 5.62 & 33.54 & 27.93 & 20.84 & 8.76 & 19.19 & 27.36 & 32.97 \\  
CoT & 16.03 & 5.58 & 33.83 & 28.13 & 21.01 & 8.53 & 18.95 & 27.77 & 38.88 \\  
VanillaRAG & 66.98 & 31.64 & 56.38 & 58.49 & 37.52 & \underline{32.3} & \underline{43.92} & 60.97 & 64.24 \\ 
RAPTOR & 66.33 & 30.98 & 58.08 & 58.72 & \underline{37.88} & 31.17 & 43.47 & OOT & OOT \\  
RAPTOR-K & 66.26 & 30.86 & 58.34 & 57.82 & 37.45 & 31.67 & 43.91 & \underline{61.24} & \underline{65.19} \\  
EraRAG & 61.40 & 29.94 & \textbf{60.79} & 58.67 & 37.89 & 26.70 & 38.93 & 61.03 & 65.22 \\  
L-LightRAG & 67.05 & \underline{39.97} & 48.05 & 53.05 & 37.38 & 30.63 & 43.66 & OOT & OOT \\  
G-LightRAG & 23.52 & 11.26 & 30.53 & 52.15 & 35.42 & 16.01 & 27.73 & OOT & OOT \\  
H-LightRAG & 47.46 & 25.91 & 38.60 & 52.86 & 36.84 & 18.73 & 29.41 & OOT & OOT \\  
HippoRAG & 55.18 & 23.35 & 46.10 & 58.88 & 31.36 & 20.00 & 32.54 & 51.89 & 56.31 \\  
HippoRAG2 & \underline{68.12} & 38.77 & 45.98 & \underline{61.66} & 33.72 & 30.23 & 39.23 & OOT & OOT \\ 
L-GraphRAG & 54.82 & 30.89 & 50.01 & 53.01 & 36.78 & 14.57 & 28.72 & OOT & OOT \\  
HiRAG & 53.18 & 30.47 & 55.83 & 54.00 & 38.09 & 26.20 & 40.70 & OOT & OOT \\ 
ArchRAG & 30.81 & 14.04 &  37.77 & 60.64 & 33.44 & 14.27 & 29.22 & \text{OOT} & \text{OOT} \\  
\midrule
\model & \textbf{72.34} & \textbf{43.51} & \underline{59.49} & \textbf{73.79 (65.41)} & \textbf{44.05 (39.44)} & \textbf{35.87} & \textbf{48.06} & \textbf{68.72} & \textbf{70.79} \\  
+Inference & 72.38 & 43.62 & 59.79 & 74.33 & 44.09 & 36.6 & 48.36 & 69.10 & 71.08 \\  
\bottomrule
\end{tabular}
}
\end{table*}

\subsubsection{Baseline}

We compare our approach with three categories of RAG baselines:
(1) LLM-only: \textbf{ZeroShot}, \textbf{Chain-of-Thought (CoT)}\cite{10.5555/3600270.3601883};
(2) Context-aware: \textbf{VanillaRAG}\cite{10.5555/3495724.3496517}, \textbf{RAPTOR}\cite{sarthi2024raptor} (GMM/K-means clustering algorithm), and \textbf{EraRAG}\cite{zhang2025eraragefficientincrementalretrieval};
(3) Relation-aware: \textbf{L-GraphRAG}\cite{edge2025localglobalGraphRAG}(local mode of Microsoft GraphRAG), \textbf{HippoRAG}\cite{10.5555/3737916.3739818}, \textbf{HippoRAG2}\cite{gutiérrez2025ragmemorynonparametriccontinual}, \textbf{HiRAG}\cite{huang2025retrievalaugmentedgenerationhierarchicalknowledge}, \textbf{ArchRAG}\cite{wang2025archragattributedcommunitybasedhierarchical} and \textbf{LightRAG}\cite{guo2025lightragsimplefastretrievalaugmented}. Variants of LightRAG, namely Local, Global, and Hybrid, are denoted as \textbf{L/G/H-LightRAG} for brevity.

\subsubsection{Implementation Details}

All static-query experiments use Llama-3.1-8B-Instruct\cite{grattafiori2024llama3herdmodels} as backbone under the vLLM\cite{kwon2023efficient} engine with greedy decoding and top-k=5 retrieval. For a fair comparison, we maintain retrieval sizes across all node types, i.e.,$\text{entity nodes} = \text{community + chunk nodes} = \text{relations} = \text{top-}k$. Embeddings are generated by BGE-M3\cite{chen2024bgem3embeddingmultilingualmultifunctionality}, a SOTA embedding model that supports both multilingual and multi-granularity retrieval, splitting text into 1,200-token chunks with 100-token overlap.
Baselines follow the framework\cite{zhou2025indepthanalysisgraphbasedrag} or official code with default hyperparameters. Runs exceeding two days are marked as \textit{OOT}.

\subsection{Static QA Performace}

To answer the question RQ1, we present the experimental results of question answering (QA) on various types of static queries, including factual accuracy, multi-hop reasoning, and reading comprehension tasks. Our method is systematically compared with a range of baseline models to assess its effectiveness across different reasoning scenarios. Furthermore, the statistics of different datasets are summarized in the Appendix \ref{sec:Statistics}.

\subsubsection{QA results}

As shown in Table~\ref{tab:static_results}, different categories of baseline methods exhibit substantial variations in static QA performance. LLM inference-only approaches (Zero-shot and CoT) perform poorly across all datasets, with accuracies significantly lower than retrieval-based methods. This indicates that relying solely on the parametric knowledge of LLMs is insufficient for factual and reasoning-intensive tasks.

Among context-level methods, VanillaRAG, RAPTOR, and EraRAG achieve comparable performance on factual accuracy and multi-hop reasoning datasets. This suggests that the advantage of high-order community structures is not always benificial, and the effectiveness largely depends on whether multi-hop content is aggregated into the same community during the offline indexing stage. On the QuALITY dataset, however, RAPTOR and EraRAG exhibit clear advantages, with EraRAG achieving the best overall performance. We attribute this to EraRAG’s retrieval strategy, which enforces the selection of one non-leaf node and four leaf nodes, thereby reducing irrelevant information and facilitating semantic understanding by the LLM. Overall, RAPTOR slightly outperforms RAPTOR-K and EraRAG, mainly due to its use of a superior but expensive clustering method, though the improvements are marginal.

For relation-level methods, the LightRAG variants demonstrate divergent strengths. L-LightRAG achieves the highest recall (39.97\%) on PopQA, showing that relation-aware retrieval can effectively support LLMs in factual accuracy tasks. In contrast, G-LightRAG and H-LightRAG perform relatively poorly, likely because global higher-level concepts are less suited for highly specific queries. ArchRAG and HiRAG, as improved versions of L-GraphRAG, refines reasoning path generation by leveraging semantically similar summary entities. HiRAG achieves better performance than L-GraphRAG in multi-hop reasoning, though still leaving room for improvement. HippoRAG and its improved version, HippoRAG2, improving retrieval performance through relations, exhibit competitive results on tasks such as MultiHop-RAG and PopQA but consistently fall short of our method in terms of overall accuracy.

Overall, \model achieves the best or near-best performance across all tasks. On PopQA, it reaches 72.34\% accuracy and 43.51\% recall, outperforming other methods by 6.2\% and 8.9\%, respectively. On the QuALITY dataset, it achieves the second-best result. In multi-hop reasoning tasks, \model attains 65.41\% accuracy on MultiHop-RAG, a 6.0\% improvement over the strongest baseline(the metric in parentheses denotes results without explicitly prompting the model to state “Insufficient Information” when possible), and shows stable advantages on MuSiQue and HotpotQA, with an average gain of approximately 11.1\%. With the addition of inference augmentation (+Inference), which add a rule in prompt to encourage the model to leverage our domain knowledge for reasoning, MuSiQue and HotpotQA further improve by 2.0\% and 0.4\%, respectively. These results demonstrate that \model consistently delivers robustness and superior effectiveness across factual QA, reading comprehension, and multi-hop reasoning tasks through a synergistic integration of relation and context.
In addition, we record in the Appendix \ref{sec:Extended_Experiments} the results of further analyses using different dense retrievers and replacing the base LLM with Qwen3-8B\cite{yang2025qwen3technicalreport}. The detailed results are shown in Table~\ref{tab:embedding_performance} and Table~\ref{tab:static_results_qwen3}. We observe that across various embedding models, \textbf{\model} consistently maintains stable and competitive performance, indicating that its advantages are not limited to a specific dense retriever. 

When replacing the base LLM with the newer model Qwen3-8B\cite{yang2025qwen3technicalreport} (using a consistent \textit{no-thinking} mode), the overall trend remains consistent with previous analyses. Notably, we observe an unexpected performance drop of context-aware methods on the MuSiQue and HotpotQA datasets after replacing the base LLM. In contrast, \model maintains competitive results, achieving up to 10\% higher accuracy than other relation-aware approaches. This demonstrates its robustness and adaptability across different model architectures and reasoning paradigms.

\begin{figure}[t!]
    \centering
    \includegraphics[width=\columnwidth]{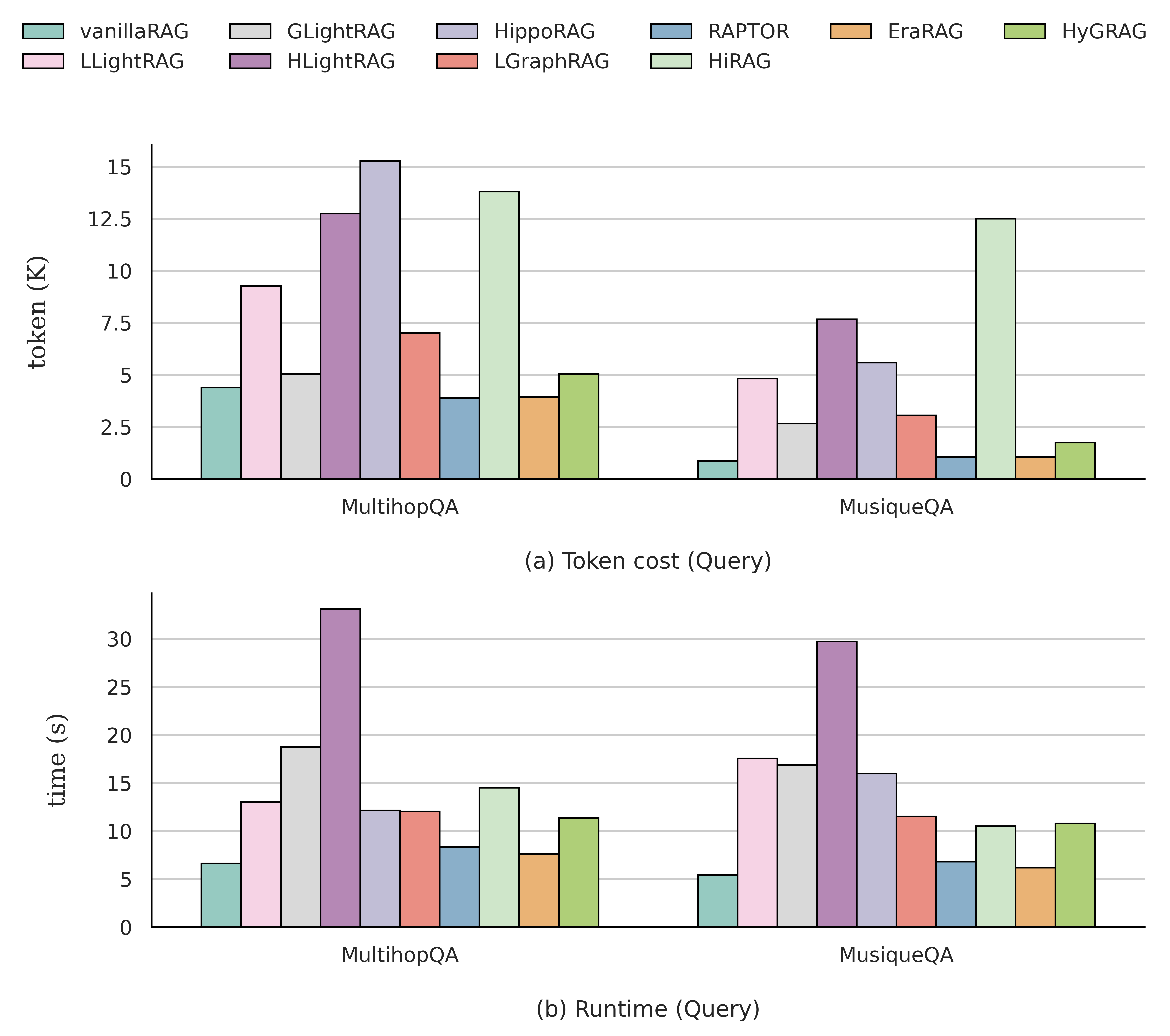}
    \setlength{\abovecaptionskip}{-0.1cm} 
    \setlength{\belowcaptionskip}{-0.4cm} 
    \caption{Comparison of query efficiency.}
    \label{fig:query_image}
\end{figure}

\subsubsection{Efficiency of \model}

To answer the question RQ2, we compare the time cost and token usage of \model with those of other baseline methods. As shown in Figure~\ref{fig:query_image}, \model demonstrates significant time and cost efficiency especially among relation-aware methods for online queries. However, compared with context-aware methods, the cost of \model is higher.

\begin{figure}[t!]
    \centering
    \includegraphics[width=\columnwidth]{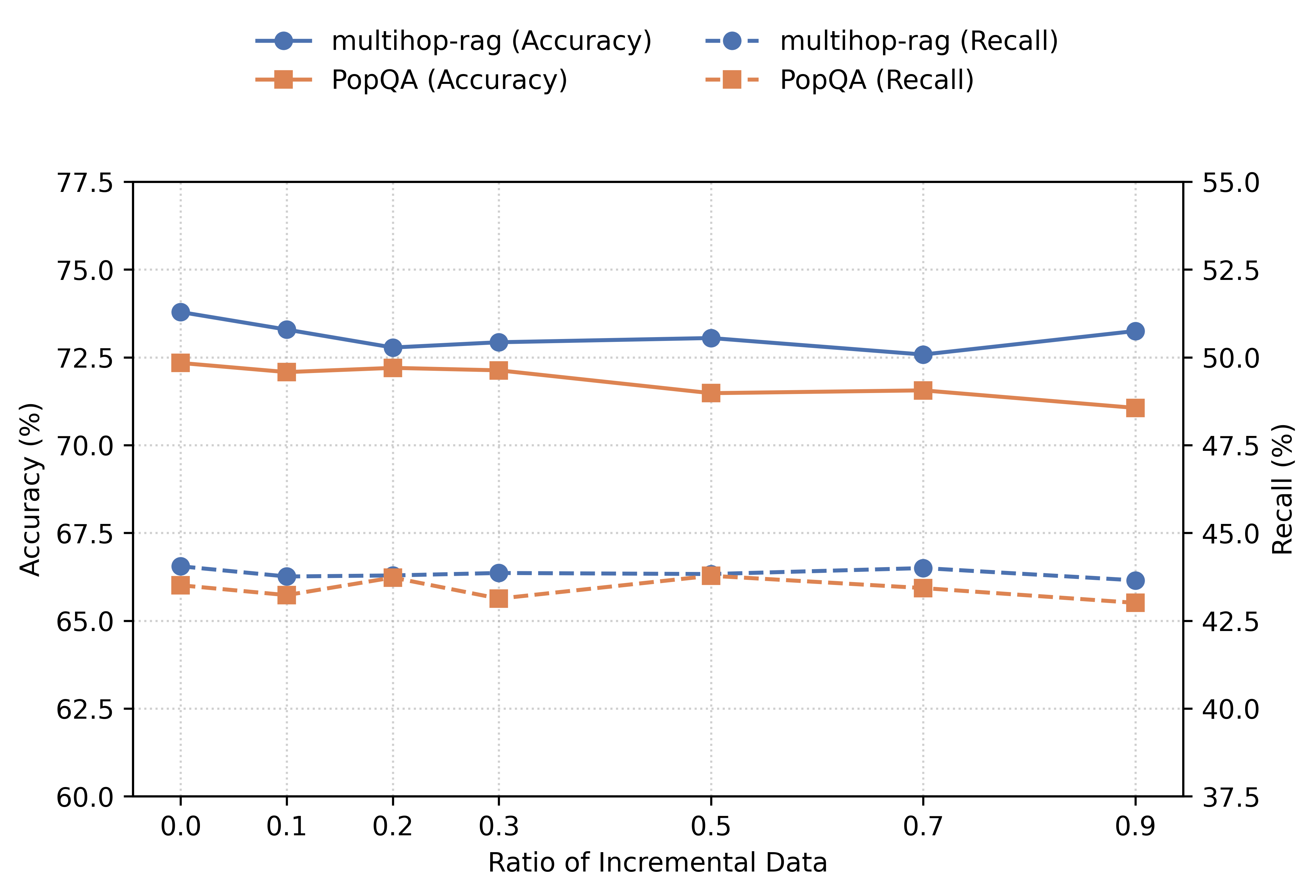}
    \setlength{\abovecaptionskip}{-0.1cm} 
    \setlength{\belowcaptionskip}{-0.3cm} 
    \caption{Corpus expansion performance.}
    \label{fig:incremental_performance}
\end{figure}

\subsubsection{Robustness to Corpus Expansion}

As RAG systems are increasingly deployed in real-world applications, they must become more adaptable to scenarios of continuous learning where the retrieval corpus keeps expanding. To answer the question RQ3, in order to evaluate the ability of \textbf{\model} to handle incremental corpus insertions, we designed an experiment that divides the initial corpus into different proportions and then incrementally inserts the remaining corpus as updates. We then tested the QA performance under varying proportions of incremental corpus, and the results are shown in Figure~\ref{fig:incremental_performance}.

\begin{figure}[t!]
    \centering
    \includegraphics[width=\columnwidth]{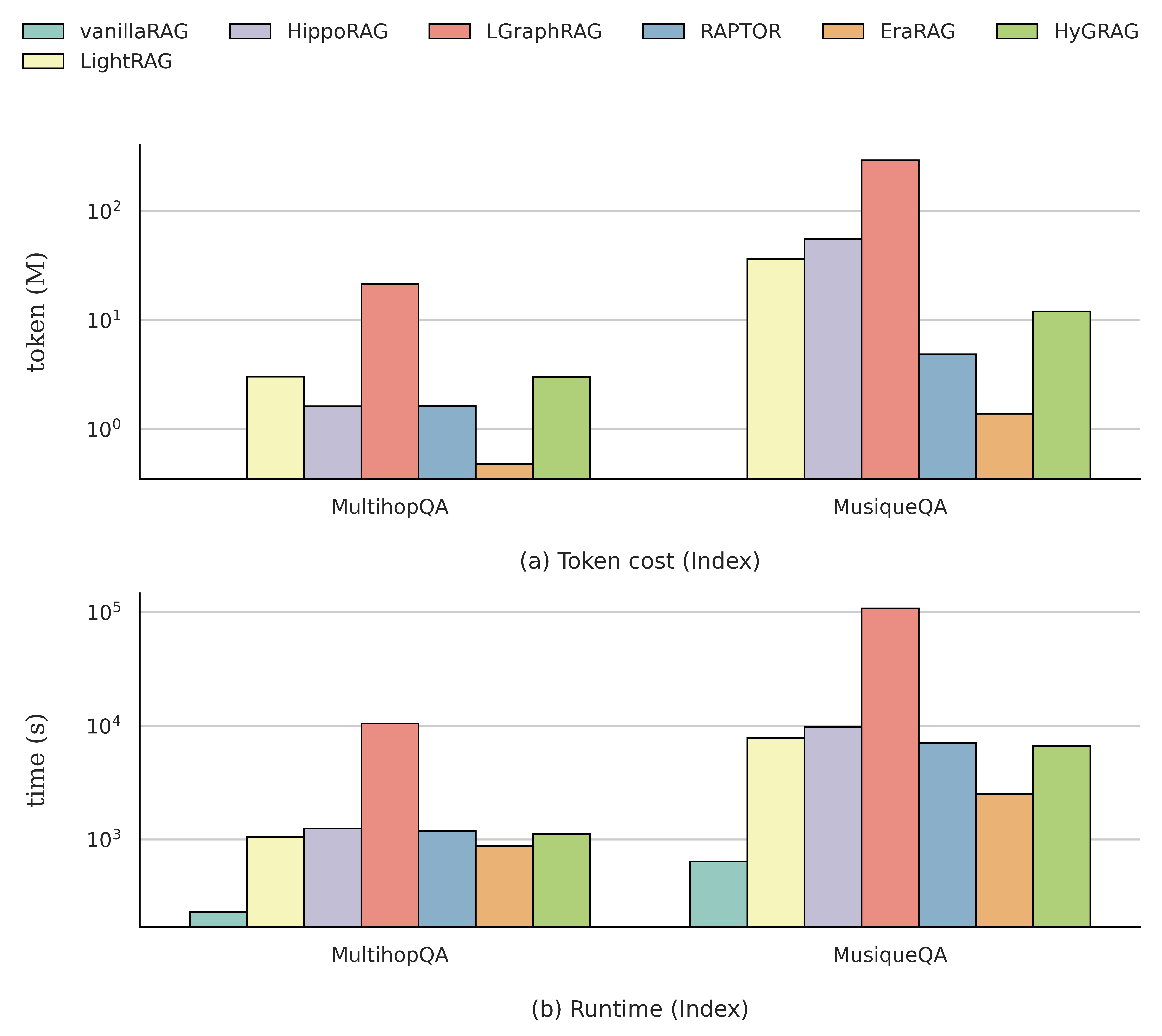}
    \setlength{\abovecaptionskip}{-0.1cm} 
    \setlength{\belowcaptionskip}{-0.4cm} 
    \caption{Corpus expansion indexing cost.}
    \label{fig:index_image}
\end{figure}

From the results, we observe that the larger the initial corpus proportion, the better the final QA performance. This is mainly because, compared to static construction, incremental corpus insertion leads to a slight decrease in community quality. However, this decrease is minor, around 1--2\%, which shows the capability of \textbf{\model} to handle continual learning scenarios effectively.
In addition, we measured the reconstruction time and token consumption required by various baseline methods when 20\% of the corpus is inserted, as shown in Figure~\ref{fig:index_image}. The results show that our proposed \textbf{\model} achieves favorable efficiency and competitive performance, indicating its strong practicality in dynamic corpus.

\subsection{Detailed Analysis}

To answer \textbf{RQ4}, we conduct detailed analyses to investigate how each component of \model cooperatively affects the overall performance of the system.

\subsubsection{Ablation Study}

Table~\ref{tab:ablation} presents the results of our ablation studies conducted on multiple datasets. To assess the individual contributions of each retrieval strategy, we progressively removed specific components and examined the resulting performance changes.

As shown in Table~\ref{tab:ablation}, removing the chunk-level structure leads to the most pronounced degradation across all datasets, highlighting its crucial role in providing fundamental contextual grounding and factual support. Similarly, eliminating entity and relation information causes a decline in accuracy, confirming the importance of relational reasoning for both factual and multi-hop question answering tasks. Interestingly, on the quality dataset, this removal yields a slight performance improvement. We attribute this to the fact that, in reading comprehension tasks—particularly for smaller LLMs—explicit relational cues may sometimes mislead the model, whereas context-aware semantic signals are more essential when the answer requires inference rather than direct textual matching.
In contrast, removing the community-level representation results in a relatively mild yet stable decrease, indicating that community structures primarily contribute to aggregating query-relevant higher-order knowledge, providing semantic summarization, and optimizing relation retrieval. Overall, these findings confirm that \model benefits substantially from the joint design of context-aware and relation-aware mechanisms.

\begin{table}[t] 
\centering
\caption{Results of ablation study.}
\label{tab:ablation}
\resizebox{0.48\textwidth}{!}{
\begin{tabular}{lccccc}
\toprule
\multirow{2}{*}{} & \multicolumn{2}{c}{\textbf{MultiHop-RAG}} & \textbf{QuALITY} & \multicolumn{2}{c}{\textbf{PopQA}} \\
\cmidrule(lr){2-3} \cmidrule(lr){4-4} \cmidrule(lr){5-6}
 & Acc & Rec & Acc & Acc & Rec \\
\midrule
\model (Full) & \textbf{65.41} & \textbf{39.44} & \underline{59.49} & \textbf{72.34} & \textbf{43.51} \\
w/o entity\&relation & 64.35 & \underline{39.08} & \textbf{60.20} & 70.80 & 41.30 \\
w/o community & \underline{65.02} & 39.01 & 58.75 & \underline{71.92} & \underline{43.18} \\
w/o chunk & 52.19 & 35.89 & 37.82 & 51.25 & 26.81 \\
vanillaRAG & 58.49 & 37.52 & 56.38 & 66.98 & 31.64 \\
\bottomrule
\end{tabular}
}
\end{table}

\begin{table}[t]
\centering
\caption{Comparative performance and token cost against combining HiRAG and RAPTOR.}
\label{tab:compare}
\resizebox{0.48\textwidth}{!}{
\begin{tabular}{lcccccc}
\toprule
\textbf{Method} & \multicolumn{3}{c}{\textbf{MultiHop-RAG}} & \multicolumn{3}{c}{\textbf{MuSiQue}} \\
\cmidrule(lr){2-4} \cmidrule(lr){5-7}
 & Acc & Rec & Tokens & Acc & Rec & Tokens \\
\midrule
HiRAG + RAPTOR & 62.36 & 37.39 & 7170.7 & 31.30 & 43.80 & 3277.2 \\
+ prompt & 70.27 & 41.37 & 7280.2 & 32.47 & 44.58 & 3768.5 \\
\model & \textbf{73.79} & \textbf{44.05} & \textbf{5030.3} & \textbf{35.87} & \textbf{48.06} & \textbf{1720.0} \\
\bottomrule
\end{tabular}
}
\end{table}

\subsubsection{Comparative Analysis}

To further examine the complementary effects of the \textit{context-aware} and \textit{relation-aware} components, we conducted comparative experiments. Specifically, we combined the retrieved chunks and communities from RAPTOR (the best-performing context-aware method) with the entities and relations from HiRAG (the modified method of MS GraphRAG in multi-hop reasoning path generation). We then compared the results with and without applying the same system prompt used in \model.

As shown in Table~\ref{tab:compare}, combining HiRAG and RAPTOR yields moderate improvements after applying the \model-style prompt but still underperforms our model in both accuracy and recall. Notably, \model achieves superior or comparable performance with substantially fewer tokens, demonstrating its efficiency in balancing reasoning depth and retrieval precision. These results demonstrate that the integrated design of \model is not a mere concatenation of relation and context representations, but rather an organic and unified framework that jointly models both aspects, enabling LLMs to more effectively leverage contextual and relational knowledge during reasoning.

\subsubsection{Case Study}

To further illustrate the qualitative advantages of our framework, we present representative examples from the PopQA and MuSiQue datasets. For each dataset, we compare \model with the best-performing baseline method. Detailed case studies of model responses are provided in the Appendix \ref{sec:case}.
The results demonstrate that \model effectively integrates relevant entities, relations, contextual information, and community-level knowledge to assist the model in constructing coherent reasoning chains and producing factually consistent answers.
For instance, in the MuSiQue multi-hop reasoning task, baseline models such as LLightRAG and VanillaRAG fail to retrieve the correct passages through semantic search or to establish the connection between Jan Klapáč’s birthplace and the target entity Prague Castle. In contrast, \model successfully leverages relational and contextual information from both the chunk and community levels to capture the latent associations and infer the correct answer.
Similarly, in a factual query from PopQA, under a setting involving multiple person entities, the model accurately identifies the target entity Paul Walker and retrieves the supporting evidence from key relation that he was born in Kilwinning.
These cases collectively illustrate that \model not only achieves unified modeling of relational and contextual knowledge but also enhances entity disambiguation and relational consistency during reasoning, thereby improving both factual accuracy and interpretability.

\section{Conclusion}
This work introduces a unified framework with context and relation-aware retrieval. \model addresses the problem that entity-centric and chunk-centric methods operate on representations anchored to original text. By clustering hybrid nodes and generating communities integrating contextual and relational information, we create summary representations beyond source documents. Our bi-level retrieval enables access to insights neither approach could achieve independently. Furthermore, \model's attachment-based update capability ensures efficient incorporation of new information. Overall, \model shows 9.7\% average improvements in multi-hop reasoning while maintaining efficiency for dynamic corpora.

\section*{Acknowledgments}
This work is supported by NSFC (No. 62322606, No. 62441605).

\newpage

\bibliographystyle{ACM-Reference-Format}
\bibliography{sample-base}

\appendix
\appendix

\begin{table}[h]
    \centering
    \small 
    \caption{Case Study on MuSiQue.}
    \label{tab:case_study1}
    \begin{tabularx}{\linewidth}{l X}
        \toprule
        \textbf{Category} & \textbf{Multi-hop reasoning} \\
        \midrule
        \textbf{Query} & 
        \textbf{What is the name of the castle in the city Jan Klapáč was born in? Answer: \underline{Prague Castle}}
        \\
        \midrule
        \textbf{LLightRAG} &
        Jan Klapáč's Birthplace: \newline
        Jan Klapáč was born in \underline{Prague}, Protectorate of Bohemia and Moravia. \newline
        
        Castle in Prague: \newline
        However, the data does not provide information about a specific castle in Prague where Jan Klapáč was born.
        \\
        \midrule
        \textbf{VanillaRAG} &
        Unfortunately, the given context does not mention Jan Klapáč. However, based on the information provided, we can see that Casimir Pulaski, a Polish general and hero of the American Revolutionary War, was born in Warsaw.
        \\
        \midrule
        \textbf{\model} &
        Analysis: \newline
        To answer the user's question, we need to find the connection between Jan Klapáč's birthplace and the castle mentioned in the context.
        
        From the **Most Relevant Entities**, we see that Prague Castle is the most relevant entity, but it's not directly related to Jan Klapáč's birthplace.
        
        However, from the **Key Relationships**, we find that Jan Klapáč was born in\underline{ Prague} (relationship 1: jan klap --[born in]--> prague).
        
        From the **Most Relevant Documents**, we see that Document 1 mentions Jan Klapáč's birthplace as Prague, Protectorate of Bohemia and Moravia.
        
        Now, let's look for a connection between Prague and a castle. From the **Key Relationships**, we find that Prague Castle is located in the Czech Republic (relationship 3: prague castle --[located in]--> czech republic).
        ...
        From the **Community Summary**, we see that \underline{Prague Castle} is a central entity associated with the architectural styles of Baroque and Mannerism...
        
        So, \underline{the name of the castle in Prague is Prague Castle}.
        \\
        \bottomrule
    \end{tabularx}
\end{table}

\section{Case Study}
\label{sec:case}

\begin{table}[h]
    \centering
    \small
    \caption{Case Study on PopQA.}
    \label{tab:case_study2}
    \begin{tabularx}{\linewidth}{l X}
        \toprule
        \textbf{Category} & \textbf{Factual accuracy} \\
        \midrule
        \textbf{Query} & 
        \textbf{In what city was Paul Walker born? Answer: \underline{Kilwinning | The Crossroads of Ayrshire}}
        \\
        \midrule
        \textbf{LLightRAG} &
        Paul Walker was born in Balwyn or Colac, Victoria, Australia. According to the data provided, both Balwyn and Colac are listed as places where Walker was born, with Balwyn being noted as a suburb in Victoria, Australia, and Colac as a town in the same region. The information suggests that Walker's birthplace is associated with these two locations, though it does not definitively state which one is the exact place of birth. This ambiguity is reflected in the sources, which mention that Walker was born in Balwyn or Colac in 1927.
        \\
        \midrule
        \textbf{RAPTOR} &
        Based on the provided information, there are multiple individuals named Paul Walker, and each has a different birthplace. 
        
        However, the most well-known Paul Walker, the actor, was born in Glendale, California, on September 12, 1973.
        
        If you are referring to Paul Baxendale-Walker, the Arctic explorer, he was born in Shrewsbury, England, on March 9, 1966.
        
        If you are referring to the other Paul Walker, the one born in Altoona, Pennsylvania, or Memphis, the information provided is not as detailed as the other two, but it can be concluded that he was born in either Altoona, Pennsylvania, or Memphis, on March 15, 1968, or in Altoona, Pennsylvania, on an unspecified date.
        \\
        \midrule
        \textbf{\model} &
        Analysis of Evidence and Reasoning:
        ...
        
        Key Relationships:
        1. **Paul Walker --[was born in]--> England**
        2. **Paul Walker --[born in]--> Scotland**
        3. **Paul Walker --[was born in]--> Shrewsbury**
        4. **Paul Walker --[born in]--> \underline{Kilwinning}**
        5. **Paul Walker --[born in]--> North Ayrshire**
        
        These relationships suggest multiple possible birthplaces for Paul Walker, including ...
        
        Documents:
        - **Document 1** states: "He was born in Shrewsbury, England on 09 March 1966."
        ...
        - **Document 5** states: "Walker was born in Franklin, Georgia, on October 22, 1935."
        
        These documents provide conflicting birthplaces for **Paul Walker**, with different cities and even different years of birth. ...
        
        Conclusion:
        Based on the **Key Relationships**, the most reliable answers are:
        - **Shrewsbury, England**
        - **\underline{Kilwinning}**
        - **North Ayrshire**
        
        These are all in **Scotland** or **England**, and they are explicitly stated in the **Key Relationships**.
        
        However, the **Documents** provide conflicting information:
        - **Document 1** says **Shrewsbury, England**
        ...
        This suggests that there may be multiple individuals named **Paul Walker**, or that the documents refer to different people with the same name....
        \\
        \bottomrule
    \end{tabularx}
\end{table}

As shown in Table~\ref{tab:case_study1}, the baselines fail to establish the correct reasoning chain required to answer the question. LLightRAG retrieves partial contextual evidence (i.e., Jan Klapáč’s birthplace) but lacks the relational linkage to the target entity, while VanillaRAG produces an entirely irrelevant response due to the absence of structured retrieval.
In contrast, \model successfully integrates entity-level, relational, and community-level information. It first identifies the entity Jan Klapáč and his birthplace (Prague), then connects this with the relation located in and the community summary mentioning Prague Castle. This multi-level reasoning enables the model to recover the implicit connection between the birthplace and the target entity, ultimately leading to the correct answer (Prague Castle).
This case illustrates that \model effectively captures and organizes cross-entity relations and contextual dependencies, allowing the LLM to perform more accurate and interpretable multi-hop reasoning.

Table~\ref{tab:case_study2} presents a case from the PopQA dataset illustrating \model’s superiority in handling entity ambiguity.
The query “In what city was Paul Walker born?” involves multiple homonymous entities. Baseline models like LLightRAG and RAPTOR confuse different individuals, producing inconsistent or irrelevant answers (e.g., Balwyn, Colac, or Glendale).
In contrast, \model combines entity–relation reasoning with chunk-level to filter out unrelated entities and consolidate consistent evidence. It correctly identifies Kilwinning and North Ayrshire—matching the gold answer—demonstrating that its relational and hierarchical design substantially improves factual accuracy and interpretability.

\begin{table*}[htbp]
\centering
\caption{Statistic of \model using Llama-3.1-8B-Instruct.}
\label{tab:graph_stats}
\begin{tabular}{lcccccc}
\toprule
\textbf{Metric} & \textbf{PopQA} & \textbf{QuALITY} & \textbf{MultiHop-RAG} & \textbf{MuSiQue} & \textbf{HotpotQA} \\
\midrule
\# of chunk nodes & 32,157 & 1,518 & 1,658 & 29,898 & 66,559 \\
\# of entity nodes & 271,776 & 33,440 & 36,884 & 224,846 & 510,457 \\
\# of community nodes & 6,403 & 758 & 843 & 5,422 & 12,109 \\
\# of relations & 379,564 & 32,938 & 32,968 & 314,145 & 758,380 \\
\# of chunk-chunk edge & 162,273 & 3,690 & 16,466 & 48,264 & 199,675 \\
\# of chunk-entity edge & 360,891 & 15,453 & 37,782 & 293,276 & 647,454 \\
\bottomrule
\end{tabular}
\end{table*}

\begin{table}[ht]
\centering 
\setlength{\abovecaptionskip}{-0.05cm} 
\setlength{\belowcaptionskip}{-0.1cm} 
\caption{Dataset Statistics}
\label{tab:dataset_statics} 
\resizebox{0.49\textwidth}{!}{
\begin{tabular}{lrrrrr}\toprule 
\textbf{Dataset} & \textbf{Tokens} & \textbf{Questions} & \textbf{of Chunks} & \textbf{Avg. Tokens} \\ \midrule
PopQA & 3,896,179 & 1,399 & 33,595(32,158) & 121.16 \\ 
QuALITY & 1,521,377 & 4,609 & 265(1,518) & 1084.25 \\ 
MultiHop-RAG & 1,424,248 & 2,556 & 609(1,658) & 920.62 \\
MuSiQue & 3,134,496 & 3,000 & 29,898 & 104.84 \\ 
HotpotQA & 8,139,124 & 3,702 & 66,581(66,555) & 122.29 \\ \bottomrule 
\end{tabular}
}
\end{table}

\begin{table}[htbp]
\centering
\caption{Hyperparameter settings used in our experiments.}
\label{tab:hyperparameters}
\resizebox{0.33\textwidth}{1.9cm}{%
\begin{tabular}{l c}
\toprule
\textbf{Parameter} & \textbf{Value} \\
\midrule
cleora\_iterations & 2 \\
lsh\_num\_hyperplanes $k$ & 16 \\
lsh\_min\_cluster\_size $S_{\min}$ & 5 \\
lsh\_max\_cluster\_size $S_{\max}$ & 50 \\
max\_hierarchy\_levels $L$ & 4 \\
community\_summary\_length & 300 \\
shared\_entity\_threshold $l$ & 3 \\
update\_th $tau_{\text{attach}}$ & 0.65 \\
\bottomrule
\end{tabular}%
}
\end{table}



\begin{algorithm}[t]
\caption{HyGRAG Hierarchical Index Construction}
\label{alg:indexing}
\begin{algorithmic}[1]
\Require Corpus $D$, Max layers $L$, Split thresholds $S_{min}, S_{max}$
\Ensure Hierarchical Index Tree $T_{index}$

\State \textbf{Phase 1: Hybrid Graph Construction}
\State $C \leftarrow \text{Chunking}(D)$ \Comment{Split documents into chunks}
\State $V^c \leftarrow C$; $E^c \leftarrow \emptyset$
\For{pair $(c_i, c_j)$ in $C$}
    \If{$|\text{Entities}(c_i) \cap \text{Entities}(c_j)| > \epsilon$}
        \State $E^c \leftarrow E^c \cup \{(c_i, c_j)\}$ \Comment{Eq. 2}
    \EndIf
\EndFor
\State $G^e \leftarrow \text{ExtractTriplets}(C)$ \Comment{LLM extracts $(h,r,t)$}
\State $E^{ce} \leftarrow \{(e, c) \mid e \in V^e, c \in C(e)\}$ \Comment{Eq. 6}
\State $G_{hybrid} \leftarrow (V^c \cup V^e, E^c \cup E^e \cup E^{ce})$

\State \textbf{Phase 2: Hierarchical Clustering}
\State $Z \leftarrow \text{Cleora}(G_{hybrid})$ \Comment{Structure-aware embeddings}
\State $\mathcal{L}_0 \leftarrow V(G_{hybrid})$
\For{$l = 1$ to $L$}
    \State $\mathcal{C}_l \leftarrow \text{LSH\_Clustering}(\mathcal{L}_{l-1}, Z)$ \Comment{Eq. 8}
    \State $\mathcal{L}_l \leftarrow \emptyset$
    \For{$bucket \in \mathcal{C}_l$}
        \State $B \leftarrow \text{AdjustSize}(bucket, S_{min}, S_{max})$
        \State $t_B \leftarrow \text{LLM}_{summ}(\{v \mid v \in B\})$ \Comment{Eq. 10}
        \State $s_B \leftarrow \text{LM}_{emb}(t_B)$ \Comment{Eq. 11}
        \State $Z(s_B) \leftarrow s_B$ \Comment{Update embeddings for next layer}
        \State $\mathcal{L}_l \leftarrow \mathcal{L}_l \cup \{s_B\}$
        \State $\text{Parent}(B) \leftarrow s_B$
    \EndFor
\EndFor
\State \Return $T_{index} = \{\mathcal{L}_0, \dots, \mathcal{L}_L\}$
\end{algorithmic}
\end{algorithm}


\begin{algorithm}[t]
\caption{Context and Relation-Aware Retrieval}
\label{alg:retrieval}
\begin{algorithmic}[1]
\Require Query $q$, Index $T_{index}$, Top-$k$ param $k$
\Ensure Generated Answer $A$

\State $q_{emb} \leftarrow \text{LM}_{emb}(q)$ \Comment{Encode query}


\State \textbf{Step 1: Context-Aware Retrieval}
\State $\mathcal{S}_{pool} \leftarrow (\bigcup_{l=1}^L \mathcal{L}_l) \cup V^c$
\State $\mathcal{N}_{ctx} \leftarrow \text{TopK}(\mathcal{S}_{pool}, q_{emb}, k)$ 
\State $\mathcal{P}_{ret} \leftarrow \{n \in \mathcal{N}_{ctx} \mid n \text{ is Community}\}$ \Comment{Separate for prompt}
\State $C_{ret} \leftarrow \{n \in \mathcal{N}_{ctx} \mid n \text{ is Chunk}\}$
\State $E^{ctx}_{ret} \leftarrow \text{TopK}(V^e, q_{emb}, k)$ \Comment{Entities retrieved separately}

\State \textbf{Step 2: Relation-Aware Retrieval}
\State $E_{expand} \leftarrow \bigcup_{C \in \mathcal{P}_{ret}} \{e \mid (e, C) \in E^{ce}\}$
\State $E_{all} \leftarrow E^{ctx}_{ret} \cup E_{expand}$ \Comment{Integrate entities}
\State $R_{all} \leftarrow \{(h,r,t) \mid h,t \in E_{all}\}$ \Comment{Extract connected triplets}
\State $R_{ret} \leftarrow \text{TopK}(R_{all}, q_{emb}, k)$ \Comment{Filter by similarity}

\State \textbf{Step 3: Generation}
\State $Ctx \leftarrow \text{Prompt}(\mathcal{P}_{ret}, C_{ret}, E^{ctx}_{ret}, R_{ret})$ \Comment{Structured Prompt}
\State $A \leftarrow \text{LLM}_{gen}(q, Ctx)$
\State \Return $A$
\end{algorithmic}
\end{algorithm}

\section{Algorithm Details}
The index construction process is illustrated in Algorithm \ref{alg:indexing}. Subsequently, the logic of the retrieval phase is detailed in Algorithm \ref{alg:retrieval}.
\model separates its operational logic into two distinct phases: hierarchical index construction and dual-aware retrieval. 
Algorithm \ref{alg:indexing} delineates the creation of a hybrid graph that bridges unstructured chunks with structured entity-relations, followed by a bottom-up community abstraction process. 
Specifically, it employs LSH-based clustering on Cleora embeddings and generates LLM summaries creating emergent knowledge.
Algorithm \ref{alg:retrieval} details the retrieval strategy, which synergizes context with relational triplets to provide a comprehensive context for the final generation. 
Specifically, it performs bi-level retrieval across hierarchy layers while expanding entity coverage through communities for comprehensive context.

\begin{table*}[ht]
\centering
\caption{Overall QA results(Accuracy and Recall) on static RAG query datasets using Qwen3-8B. ``OOT'' denotes results that could not be obtained within two days.}
\label{tab:static_results_qwen3}
\resizebox{0.8\textwidth}{!}{
\begin{tabular}{lccccccccc} 
\toprule
\textbf{Method} & \multicolumn{2}{c}{\textbf{Factual Accuracy}} & \multicolumn{3}{c}{\textbf{Reading Comprehension}} & \multicolumn{4}{c}{\textbf{Multi-Hop Reasoning}} \\
\cmidrule(lr){2-3} \cmidrule(lr){4-6} \cmidrule(lr){7-10}
& \multicolumn{2}{c}{PopQA} & \multicolumn{1}{c}{QuALITY} & \multicolumn{2}{c}{MultiHop-RAG} & \multicolumn{2}{c}{MuSiQue} & \multicolumn{2}{c}{HotpotQA} \\ 
\cmidrule(lr){2-3} \cmidrule(lr){4-4} \cmidrule(lr){5-6} \cmidrule(lr){7-8} \cmidrule(lr){9-10}
& \multicolumn{1}{c}{Accuracy} & \multicolumn{1}{c}{Recall} & \multicolumn{1}{c}{Accuracy} & \multicolumn{1}{c}{Accuracy} & \multicolumn{1}{c}{Recall} & \multicolumn{1}{c}{Accuracy} & \multicolumn{1}{c}{Recall} & \multicolumn{1}{c}{Accuracy} & \multicolumn{1}{c}{Recall} \\
\midrule
Zero-shot & 25.23 & 9.83 & 34.78 & 47.10 & 27.33 & 8.13 & 22.38 & 30.01 & 36.62 \\
CoT & 30.52 & 14.75 & 36.06 & 60.02 & 31.37 & 10.50 & 27.97 & 32.90 & 41.15 \\
VanillaRAG & 63.83 & 35.00 & 59.28 & 61.93 & 37.40 & 23.60 & 35.26 & 51.54 & 55.15 \\
RAPTOR & 63.18 & 31.05 & \underline{60.87} & 60.13 & 36.66 & 23.67 & 35.37 & \underline{54.70} & \underline{59.07} \\
EraRAG & \text{OOT} & \text{OOT} & \textbf{61.78} & 61.78 & 36.85 & \text{OOT} & \text{OOT} & \text{OOT} & \text{OOT} \\
L-LightRAG & \underline{71.05} & \underline{44.37} & 58.41 & 55.52 & \underline{38.25} & \underline{33.10} & \underline{47.52} & \text{OOT} & \text{OOT} \\
HippoRAG & 63.97 & 33.49 & 56.65 & 61.89 & 33.54 & 24.90 & 38.88 & \text{OOT} & \text{OOT} \\
HippoRAG2 & 67.12 & 35.22 & 42.53 & \underline{63.97} & 36.24 & 25.87 & 37.08 & \text{OOT} & \text{OOT} \\
L-GraphRAG & 58.32 & 33.24 & 58.75 & 55.20 & 38.04 & 18.53 & 32.79 & \text{OOT} & \text{OOT} \\
\midrule
\model & \textbf{73.84} & \textbf{46.17} & 60.67 & \textbf{73.87 (67.57)} & \textbf{45.09 (38.87)} & \textbf{36.67} & \textbf{49.33} & \textbf{72.18} & \textbf{72.67} \\
\bottomrule
\end{tabular}
}
\end{table*}

\section{Experiment Details}
\label{sec:Statistics}
We show the hybrid graph statistics, dataset statistics and Hyperparameter settings in Table~\ref{tab:graph_stats}, Table~\ref{tab:dataset_statics} and Table~\ref{tab:hyperparameters}. Exploring GNN~\cite{DBLP:journals/corr/KipfW16, ijcai2022p310, sun2024fine, Sun2025HandlingFH, sun2025mlnc} to learn more expressive representations over heterogeneous structures is a future work.
Regarding efficiency (on Musique), \model’s offline stage (23,042s) is 21\% faster than the efficient GraphRAG representative (LightRAG). Online stage (10.7s) is 7\% faster than all graph-based baselines. During the process, it uses 9.97GB memory and 2.50GB GPU (for embedding-model only). Clustering takes 31s (GMM-clustering in RAPTOR took 10,965s). Summarization stage requires ~1 hour, about 15\% of overall indexing time. 
Results show \model is efficient in both offline and online phases.
Regarding hallucinations in LLM-generated community summaries, due to inherent limitations of LLM paradigm, using LLMs inevitably introduces certain probability of hallucinations (errors). However this issue can be mitigated as LLMs become more powerful. Our framework allows replacing the LLM component, and even lightweight models still improve performance. We manually evaluate 100 sampled summaries on Musique and found 17\% hallucination rate, mostly due to added external knowledge. After switching to a stronger model (gpt-4o-mini) the rate dropped to 7\%, indicating improved summary quality.

\begin{table}[h]
\caption{Performance comparison of different embedding models under Vanilla and \model settings.}
\begin{tabular}{lccccc}
\toprule
 Dense Retriever &  & \multicolumn{2}{c}{Vanilla} & \multicolumn{2}{c}{\model} \\
\cmidrule(lr){3-4} \cmidrule(lr){5-6}
 &  & Accuracy & Recall & Accuracy & Recall \\
\midrule
bge-m3                         &  & 58.49    & 37.52  & 65.41          & 39.44  \\
Qwen3-E-0.6B           &  & 57.98    & 37.45  & 65.49          & 39.39  \\
text-embedding-v3              &  & 59.31    & 38.14  & 66.00          & 39.74  \\
m-e5-large-instruct &  & 58.76    & 37.83  & 63.65          & 39.06  \\
\bottomrule
\end{tabular}
\label{tab:embedding_performance}
\end{table}

\section{Extended Experiments}
\label{sec:Extended_Experiments}
\subsection{Robustness in Dense Retrievers}

We experimented with a range of SOTA embedding models. As shown in Table~\ref{tab:embedding_performance}, \model consistently outperforms the corresponding vanilla dense retrieval baseline across all embedding models. Moreover, the performance of \model improves progressively as the quality of the underlying embedding model increases, demonstrating its strong scalability and sensitivity to embedding fidelity. In our experiments, we specifically evaluated \model using Qwen3-Embedding-0.6B\cite{zhang2025qwen3embeddingadvancingtext}, text-embedding-v3\cite{zhang2025qwen3embeddingadvancingtext}, and multilingual-e5-large-instruct\cite{wang2024multilinguale5textembeddings} as the underlying dense retrievers.
This demonstrates that \model's retrieval framework effectively amplifies embedding quality, creating synergistic gains beyond vanilla dense retrieval approaches.

\subsection{Robustness in Base LLM}
\label{sec:Qwen3-8B}
We present the QA performance of various methods using the base LLM Qwen3-8B, as summarized in Table~\ref{tab:static_results_qwen3}. When employing Qwen3-8B for both indexing and question answering, \model consistently achieves competitive accuracy and recall across most datasets, maintaining leading performance on MuSiQue and HotpotQA, with accuracy gains up to 21.36\% over other methods.

For context-level methods, VanillaRAG, RAPTOR, and EraRAG show slight improvements compared to their performance with the previous base LLM. However, they exhibit noticeable drops on the MuSiQue and HotpotQA datasets, indicating that their performance is sensitive to changes in the base LLM. Similarly, the relation-level method HippoRAG2 also shows performance declines on these datasets, and its results on QuALITY remain suboptimal.

In contrast, \model maintains stable and superior performance across all tasks, demonstrating robustness and adaptability to different LLM architectures. This suggests that \model’s synergistic integration of relation- and context-level information provides resilience against base model changes, while other approaches are more sensitive to variations in the underlying LLM.

\section{Prompts Used for \model}

The Answer Generation Prompt is designed to synthesize a comprehensive and evidence-grounded answer to a user's query from a structured, multi-source context. The process begins by presenting the model with a rich contextual payload retrieved from a hierarchical knowledge system, which includes: hierarchical community summaries with similarity scores, a ranked list of the most relevant entities, key entity relationships structured as triplets, and the content of the most relevant source documents. Subsequently, the prompt provides a set of explicit operational rules to constrain the generation process. The framework mandates that the model must report any informational gaps by stating "Insufficient information" and is forbidden from fabricating information not present in the context, ensuring a high degree of fidelity to the source material.

The Community Summarization Prompt outlines a framework for distilling a collection of content, referred to as a knowledge community, into a structured and semantically dense summary. The prompt's primary objective is to generate a summary specifically tailored for the downstream task of creating high-quality semantic embeddings. It guides the model by defining a clear, four-part structure for the output, requiring the summary to comprehensively cover: 1) key themes and topics, 2) important entities and their roles, 3) relationships and connections between these entities, and 4) the overall context and significance of the information. By specifying both a word limit and the explicit need for rich semantic content, the prompt ensures the generation of a concise yet information-rich text that captures the core essence of the knowledge community.

The detailed prompt are provided in the code repository \url{https://github.com/zjunet/HyGRAG}.

\end{document}